\documentclass[runningheads]{llncs}
\usepackage{graphicx}
\usepackage{booktabs}
\usepackage{multirow}
\usepackage[normalem]{ulem}
\usepackage[sectionbib, numbers]{natbib}
\useunder{\uline}{\ul}{}

\usepackage{amsmath}
\usepackage{framed}
\usepackage{mdframed}
\usepackage{lipsum}

\usepackage{hyperref}

\usepackage{footnote}
\usepackage{enumitem}
\usepackage{longtable}
\usepackage{color}
\usepackage[table,dvipsnames]{xcolor}
\usepackage{adjustbox}
\usepackage{marvosym}

\usepackage{diagbox}
\usepackage{array}
\newcolumntype{P}[1]{>{\centering\arraybackslash}p{#1}}
\usepackage{makecell}

\begin{document}
\title{Does Informativeness Matter? Active Learning for Educational Dialogue Act Classification}
\titlerunning{Active Learning for Educational Dialogue Act Classification}
% Does Informativeness Matter? Active Learning for Educational Dialogue Act Classification
% Beyond Random Sampling: Uncovering Informative Dialogue Acts through Active Learning
% Uncovering Informative Gems: Active Learning for Educational Dialogue Act Classification


\author{Wei Tan\inst{1}, 
Jionghao Lin\inst{1, 2}\textsuperscript{(\Letter)}, 
David Lang\inst{3},
Guanliang Chen\inst{1},
Dragan Gasevic\inst{1},
Lan Du\inst{1},
and Wray Buntine\inst{4,1}
}

\authorrunning{W. Tan et al.}

\institute{Monash University, Clayton, Australia \\
\email{\{wei.tan2, jionghao.lin1, lan.du, guanliang.chen, dragan.gasevic\}@monash.edu} \and
Carnegie Mellon University, Pittsburgh, USA \and
Stanford University, Stanford, USA \\
\email{dnlang86@stanford.edu} \and
VinUniversity, Hanoi, Vietnam \\
\email{wray.b@vinuni.edu.vn}\\
}
\maketitle              % typeset the header of the contribution
%
% Understanding the sample informativeness during the classifier training process

\vspace{-3mm}

\begin{abstract}
Dialogue Acts (DAs) can be used to explain what expert tutors do and what students know during the tutoring process. Most empirical studies adopt the random sampling method to obtain sentence samples for manual annotation of DAs, which are then used to train DA classifiers. However, these studies have paid little attention to sample informativeness, which can reflect the information quantity of the selected samples and inform the extent to which a classifier can learn patterns. Notably, the informativeness level may vary among the samples and the classifier might only need a small amount of low informative samples to learn the patterns. Random sampling may overlook sample informativeness, which consumes human labelling costs and contributes less to training the classifiers. As an alternative, researchers suggest employing statistical sampling methods of Active Learning (AL) to identify the informative samples for training the classifiers. However, the use of AL methods in educational DA classification tasks is under-explored. In this paper, we examine the informativeness of annotated sentence samples. Then, the study investigates how the AL methods can select informative samples to support DA classifiers in the AL sampling process. The results reveal that most annotated sentences present low informativeness in the training dataset and the patterns of these sentences can be easily captured by the DA classifier. We also demonstrate how AL methods can reduce the cost of manual annotation in the AL sampling process.

\keywords{Informativeness \and  Active Learning \and 
Dialogue Act Classification \and
Large Language Models \and Intelligent Tutoring Systems.}
\end{abstract}
% \vspace{-9mm}
%
%
%

\section{Introduction}
Traditional one-on-one tutoring involves human participants (\textit{e.g.,} a course tutor and a student), which has been widely acknowledged as an effective form of instruction \cite{rus2017dialogue, Vail2014IdentifyingEM}. Understanding what expert tutors do and students know during the tutoring process is a significant research topic in the field of artificial intelligence in education \citep{du2016modelling}, which can contribute to the design of dialogue-based intelligent tutoring systems (\textit{e.g.,} AutoTutor \citep{nye2014autotutor}) and the practice of human tutoring \cite{lin2022good, Vail2014IdentifyingEM}. A typical method used to understand the tutoring process is to map the sentences from tutors and students onto dialogue acts (DAs) which can manifest the intention behind sentences \cite{Vail2014IdentifyingEM, rus2017dialogue}. For example, a tutor's sentence (\textit{e.g.,} ``\textit{No, it is incorrect!}'') can be coded as the  \texttt{Negative Feedback} DA. The mapping process often needs a pre-defined DA scheme developed by educational experts, and the DA scheme is used to annotate the sentences with the DAs  \citep{rus2017analysis, rus2017dialogue}. However, manually annotating the DAs for millions of tutoring dialogues is impractical for conducting research related to educational dialogue since the annotating process is often time-consuming and costly \cite{nye2015automated}. Therefore, researchers typically annotate a small amount of sentences from tutoring dialogues and further use the annotated sentences to train machine learning classifiers to automate the annotation process for DAs \citep{rus2017dialogue, lin2022good}.

Though previous works demonstrated the potential of automating the DA classification \cite{nye2015automated, rus2017analysis, lin2022good, d2010mining, ezen2015understanding}, little attention has been given to investigate the extent to which the DA classifiers can learn the patterns from the annotated sentences, which is important to build a robust classifier to facilitate tutoring dialogue analysis. The ability of the DA classifiers that learn the relation between the model inputs (\textit{e.g.,} sentences) and outputs (\textit{e.g.,} DAs) is defined as learnability \cite{swayamdipta2020dataset, karamcheti2021mind}. To improve the learnability of a DA classifier, it is necessary to train the DA classifier on the instances that can help the classifier capture the underlying patterns and further improve the generalization of the DA classifier on the full dataset \cite{karamcheti2021mind}. These instances are considered highly informative samples in the dataset \cite{karamcheti2021mind}. It should be noted that the informativeness level might be different among the samples, and the classifier can learn the patterns of the low informative samples with a small amount of training data \cite{swayamdipta2020dataset, karamcheti2021mind}. Therefore, annotating more high informative samples and less low informative samples could further save human annotation costs. However, few works investigated sample informativeness on the task of educational DA classification, which motivated us to focus on the \textbf{first research goal}, \textit{i.e.,} \textit{investigate the sample informativeness levels of annotated DAs}. Additionally, the previous works annotated the sentences by randomly sampling from the archived tutoring dialogue corpora \cite{rus2017analysis, nye2015automated, rus2017dialogue, lin2022good}, which might not provide sufficient high informative samples for training the DA classifier. We argue that blindly annotating the dialogue sentences to train the DA classifier might consume excessive human annotation costs and not enhance the classifier's learnability on the full dataset \cite{settles_active}. Instead, a promising solution is to use the statistical active learning (AL) methods, which aim to select highly informative samples to train the classifier \cite{settles_active}. To provide details about the efficacy of AL methods, the \textbf{second research goal} of the current study was to \textit{investigate the extent to which statistical AL methods support the DA classifier training on the DAs with different informativeness levels.}

Our results showed that most annotated instances from the training samples presented low informativeness, which indicated that the patterns of these samples could be sufficiently captured by our DA classifier. Additionally, the DA classifier needs to train on more high informativeness samples to improve the classification performance on the unseen dataset, which further requires support by statistical AL methods. We compared the state-of-the-art AL method (\textit{i.e.,} CoreMSE \cite{tan2021diversity}) with other commonly-used AL methods (\textit{i.e.,} Least Confidence and Maximum Entropy) and random baseline. We found that the CoreMSE AL method could select more high informative instances for the DA classifier at the early stage of the classifier training process and gradually reduced the selection of the low informative instances, which could alleviate the cost of manual annotation.

\vspace{-3mm}

\section{Related Work}

\vspace{-1mm}
\subsection{Educational Dialogue Act Classification}
\label{rw_part_1}
\vspace{-1mm}

To automate classifying educational DAs, previous studies in the educational domain have employed machine learning models to train on annotated sentences \cite{nye2015automated, boyer2010dialogue, samei2014context, rus2017analysis, lin2022good}. For example, Boyer \textit{et al.} \cite{boyer2010dialogue} employed a Logistic Regression model to train on linguistic features (\textit{e.g.,} N-grams) from the annotated sentences in 48 programming tutoring sessions. They \cite{boyer2010dialogue} achieved 63\% accuracy on classifying thirteen DAs. Later on, Samei \textit{et al.} \cite{samei2014context} used Decision Trees and Naive Bayes models to train on linguistic features and contextual features (\textit{e.g.,} the speaker's role of previous sentences) from 210 annotated tutoring sentences on the science-related topic. Their work \cite{samei2014context} achieved the accuracy of 56\% in classifying seven educational DAs. These previous works trained DA classifiers by using the sentences which were randomly sampled from the larger tutoring dialogue corpora. Though these works demonstrated the feasibility of automating DA classification \cite{nye2015automated, boyer2010dialogue, samei2014context, rus2017analysis, lin2022good}, it still remains largely unknown whether the DA classifier learned the patterns sufficiently from the randomly selected sentences, which is important for building robust and reliable DA classifiers.

\vspace{-3mm}

\subsection{Sample Informativeness}
\vspace{-1mm}
\label{rw_part2}
Training a DA classifier on a sufficient amount of annotated sentences can help the classifier achieve satisfactory classification performance. However, manual annotation is time-consuming and expensive \citep{lin2022good, nye2015automated, rus2017analysis}. To mitigate the human annotation cost and also help the classifier achieve satisfied classification performance, Du \textit{et al.} \cite{du2015exploring} suggested annotating the most informative samples, which can reduce the generalization error and uncertainty of the DA classifier on the unseen data. A recent work \cite{swayamdipta2020dataset} proposed the \textit{Data Maps} framework which used \textbf{Confidence}, \textbf{Variability}, and \textbf{Correctness} to measure the informativeness. \textbf{Confidence} denotes the averaged probability of the classifier's prediction on the correct labels across all training epochs; \textbf{Variability} denotes the spread of probability across the training epochs, which capture the uncertainty of the classifier; \textbf{Correctness} denotes the fraction of the classifier correctly predicted labels over all training epochs \cite{swayamdipta2020dataset}. Building upon the work introduced in \cite{swayamdipta2020dataset}, Karamcheti \textit{et al.} \cite{karamcheti2021mind} further used the \textbf{Correctness} to categorize the samples into four groups: \textit{Easy}, \textit{Medium}, \textit{Hard}, and \textit{Impossible} \cite{swayamdipta2020dataset} and these groups indicated the extent to which the classifier can learn the patterns from the annotated instances. Inspired by \cite{du2015exploring, swayamdipta2020dataset, karamcheti2021mind}, we argue that when scrutinizing the sampling process for training the DA classifier, if samples are randomly selected, there will be redundant for overly sampling the \textit{Easy} samples and insufficient for sampling high informative samples, which might consume the human annotation budget and lead to poor generalizability of the DA classifier. As a remedy, it is important to select the most suitable samples for training the classifier. AL can offer promising methods to select the most suitable samples.

\vspace{-3mm}

\subsection{Statistical Active Learning}
\label{rw_part3}

Recent studies on educational DA classification \cite{lin2022good, nye2015automated} have agreed that the high demand for the annotated dataset was still an issue for the DA classification task. To alleviate this issue, a promising solution is to use AL methods which can select the high informative samples from the unlabeled pool and send them to Oracle (\textit{e.g.,} human annotator) for annotation \cite{settles_active}. Traditionally, there are three typical scenarios of AL methods: 1) \textit{membership query synthesis}, which focuses on generating artificial data-point for annotation rather than sampling the data-point from the real-world data distribution, 2) \textit{stream-based sampling}, which focuses on scanning through a sequential stream of non-annotated data-points and make sampling query decision individually, and 3) \textit{pool-based sampling} which focuses on selecting the most informative samples from the non-annotated data pool and send them to the oracle for annotation \cite{settles_active}.

As the annotated DAs were available in our study and the dataset was not collected in a sequential stream, we consider our study fits well with the \textit{pool-based sampling} scenario. The pool-based AL methods can both reduce the computational cost of model training and maintain the performance of the model trained on the annotated dataset \cite{settles_active}. Many studies employed the pool-based AL methods (\textit{e.g.,} Least Confidence \cite{sha-etal-2022-bigger, hastings2018active} and Maximum Entropy \cite{karumbaiah2021using}) on the educational tasks (\textit{e.g.,} student essay classification \cite{hastings2018active}, and educational forum post classification \cite{sha-etal-2022-bigger}) and their results have demonstrated the promise of AL methods on alleviating the demand for annotated datasets. However, it still remains largely unknown about the extent to which AL methods can select the informative samples to support the automatic classification of educational DA.

\vspace{-3mm}

\section{Methods}
\vspace{-2mm}
\subsection{Dataset}
\label{sec:methods:subsec:dataset}
\vspace{-1mm}

The current study obtained ethics approval from the Monash University Human Research Ethics Committee under application number 26156. The dataset used in our study was provided by an educational technology company that operated online tutoring services and collected the data from tutors and students along with informed consent allowing the use of the de-identified dataset for research. The dataset contained detailed records of the tutoring process where tutors and students collaboratively solve various problems (the subjects including mathematics, physics, and chemistry) via textual message. In total, our dataset contained 3,626 utterances (2,156 tutor utterances and 1,470 student utterances) from 50 tutorial dialogue sessions. The average number of utterances per tutorial dialogue session was 72.52 (\textit{min} = 11, \textit{max} = 325); tutors averaged 43.12 utterances per session (\textit{min} = 5, \textit{max}  = 183), and students 29.40 utterances per session (\textit{min} = 4, \textit{max} = 142). We provided a sample dialogue in the digital appendix via \url{https://github.com/jionghaolin/INFO}.

\vspace{-3mm}

\subsection{Educational Dialogue Act Scheme and Annotation}
\label{sec:methods:subsec:da_scheme}

Identifying the DAs in tutorial dialogues often relies on a pre-defined educational DA coding scheme \cite{HENNESSY201616}. By examining the existing literature, we employed the DA scheme introduced in \cite{Vail2014IdentifyingEM} whose effectiveness in analyzing online one-on-one tutoring has been documented in many previous studies (\textit{e.g.,} \cite{ezen2015classifying, lin2022good, vail2016predicting}). The DA scheme developed in \cite{Vail2014IdentifyingEM} characterizes the DAs into a two-level structure. To discover more fine-grained information from tutor-student dialogue, in our study, we decided to annotate the tutoring dialogues by using the second-level DA scheme. Notably, some utterances in dialogues contained multiple sentences, and different sentences can indicate different DAs. To address this concern, Vail and Boyer \cite{Vail2014IdentifyingEM} suggested partitioning the utterances into multiple sentences and annotating each sentence with a DA. After the utterance partition, we then removed the sentences which only presented meaningless symbols or emoji. Lastly, we recruited two human coders to annotate DAs, and we obtained Cohen's $\kappa$ score of 0.77 for the annotation. The annotations achieved a substantial agreement between the two coders, and we recruited a third educational expert to resolve the inconsistent cases. The full DA scheme can be found in a digital appendix via \url{https://github.com/jionghaolin/INFO}, which contains 31 DAs. Due to the space limit, we only presented students' DAs in Table \ref{tab:dialogue_act_scheme}.

\vspace{-3mm}

\begin{table*}[!htb]
\centering
\caption{The DA scheme for annotating student DAs. The DAs were sorted based on their frequency (\textit{i.e.,} the column of \textbf{Freq.}) in the annotated dataset.}
\vspace{-2mm}
\label{tab:dialogue_act_scheme}
\resizebox{0.75\textwidth}{!}{%
\renewcommand{\arraystretch}{1.15}
\begin{tabular}{p{.4\textwidth}p{.5\textwidth}P{.1\textwidth}}
\toprule
\textbf{Dialogue Acts (DAs)} & \textbf{Sample Sentences} & \textbf{Freq.} \rule{0pt}{2.6ex}\rule[-2.2ex]{0pt}{0pt} \\ \toprule
% General Positive Feedback & \multirow{14}{*}{\textbf{T}} & \textit{``Well done!''} & 9.46\% \\
% Information &  & \textit{``The last card also has to be a 7.''} & 8.42\% \\
% Probing Question &  & \textit{``Can you quickly simplify 6/24?''} & 8.07\% \\
% Factual Question &  & \textit{``How did you find an irrational number?''} & 3.37\% \\
% Operational Question &  & \textit{``Do you have any other questions for me?''} & 3.21\% \\
% Directive &  & \textit{``Take a look at what's circled in yellow.''} & 3.14\% \\
% Hint by Image &  & \textit{{[}Image{]}} & 2.34\% \\
% Reassurance &  & \textit{``No problem, we will do it together.''} & 2.26\% \\
% Elaborated Positive Feedback &  & \multicolumn{1}{l}{``Yep, 5 is right!''} & 2.08\% \\
% Lukewarm Feedback &  & \textit{``You are almost there!''} & 1.49\% \\
% Evaluation Question &  & \textit{``Does that make sense?''} & 1.44\% \\
% Negative Feedback &  & \textit{``Hmmm, not quite.''} & 1.42\% \\
% Open Question &  & \textit{``What is the next step?''} & 0.97\% \\
% Ready Question &  & \textit{``Ready for the Question 18?''} & 0.09\% \\ \toprule
\texttt{Confirmation Question} &  \textit{``So that'd be 5?''} & 4.93\% \\
\texttt{Request Feedback by Image} &  \textit{{[}Image{]}} & 4.34\% \\
\texttt{Understanding} &  \textit{``Oh, I get it''} & 1.46\% \\
\texttt{Direction Question} &  \textit{``Okay what do we do next?''} & 1.20\% \\
\texttt{Information Question} &  \textit{``Isn't there a formula to find the nth term?''} & 1.06\% \\
\texttt{Not Understanding} &  \textit{``I don't know.''} & 0.24\% \\
\texttt{Ready Answer} &  \textit{``Yep, ready to go.''} & 0.07\% \\
% Acknowledge & \multirow{10}{*}{\textbf{T\&S}} & \textit{``Ok!''} & 7.15\% \\
% Yes-No Answer &  & \textit{``No, I don't have any progress''} & 6.02\% \\
% Extra Domain Other &  & \textit{``Ok, I'll load again.''} & 5.94\% \\
% WH Answer &  & \textit{``The 4 point questions''} & 5.92\% \\
% Observation &  & \textit{``We initially had these 5 terms.''} & 4.36\% \\
% Greeting &  & \textit{``Hello!''} & 3.54\% \\
% Explanation &  & \textit{``That's what we've been doing''} & 3.16\% \\
% Extra Domain  Question &  & \textit{``Did you cover this in your class?''} & 1.20\% \\
% Correction &  & \textit{``Oops sorry 100 degrees''} & 0.85\% \\
% Extra Domain Answer & & \textit{``I don't know how the payments work.''} & 0.75\% \\ 
\bottomrule
\end{tabular}%
}
\vspace{-5mm}
\end{table*}
\vspace{-5mm}

\subsection{Identifying Sample Informativeness via Data Maps}
\label{data_maps}
To identify the informativeness of annotated instances, we need to train the annotated dataset on a classifier. Building upon our previous work \cite{tlt_lin_2022}, we used ELECTRA \cite{clark2019electra} as the backbone model for classifying 31 DAs, which is effective in capturing nuanced relationships between sentences and the DAs. The dataset (50 dialogue sessions) was randomly split to \textit{training} (40 sessions) and \textit{testing set} (10 sessions) in the ratio of \textit{80\%:20\%} for training the classifier. The classifier achieved accuracy of 0.77 and F1 score of 0.76 on the testing set. Then, we applied the \textit{Data Maps}\footnote{\url{https://github.com/allenai/cartography}} to the DA classifier to analyze the behaviour of the classifier on learning individual instance during the training process. Following the notation of \textit{Data Maps} in \cite{swayamdipta2020dataset}, the training dataset denotes $Dataset = \{(x, y^{*})_i\}^N_{i=1}$ across $E$ epochs where the $N$ denotes the size of the dataset, $i$th instance is composed of the pair of the observation of $x_i$ and true label $y^{*}_i$ in the dataset. \textbf{Confidence} ($\hat{\mu_i}$) was calculated by $\hat{\mu_i} = \frac{1}{E} \sum_{e = 1}^E p_{\theta^{(e)}}(y^{*}_i | x_i)$ where $p_{\theta^{(e)}}$ is the model's probability with the classifier's parameters $\theta^{(e)}$ at the end of the $e^{th}$ epoch. \textbf{Variability} ($\hat{\sigma_i}$) was calculated by the variance of $p_{\theta^{(e)}}(y^{*}_i | x_i)$ across epochs: $\hat{\sigma_i} =  \sqrt{\frac{\sum_{e=1}^E (p_{\theta^{(e)}}  (y^{*}_i | x_i) - \hat{\mu_i})^2} {E}}$. \textbf{Correctness} ($Cor$) was the fraction of the classifier correctly annotated instance $x_i$ over all epochs. A recent study \cite{karamcheti2021mind} further categorized \textbf{Correctness} into \textit{Easy} ($Cor \geq 0.75$), \textit{Medium} ($0.75 > Cor \geq 0.5$), \textit{Hard} ($0.5 > Cor \geq 0.25$), and \textit{Impossible} ($0.25 > Cor \geq 0$). In line with the studies \cite{swayamdipta2020dataset, karamcheti2021mind}, we mapped the training instances along two axes: the $y-axis$ indicates the confidence and the $x-axis$ indicates the variability of the samples. The colour of instances indicates \textbf{Correctness}.

% \vspace{-3mm}

\subsection{Active Learning Selection Strategies}
\label{method_sec2}

We aimed to adopt a set of AL methods to examine the extent to which AL methods support the DA classifier training on the DAs with different informativeness levels. We employed the state-of-the-art AL method (CoreMSE \citep{tan2021diversity}) to compare with two commonly used AL methods (\textit{i.e.,} Least Confidence \cite{settles_active} and Maximum Entropy \cite{yang2016active}) and the random baseline. Following the algorithms in the papers, we re-implemented them in our study. \textbf{Random Baseline} is a sampling strategy that samples the number of instances uniformly at random from the unlabeled pool \cite{settles_active}. \textbf{Maximum Entropy} is an uncertainty-based method that chooses the instances with the highest entropy scores of the predictive distribution from the unlabeled pool \cite{yang2016active}. \textbf{Least Confidence} is another uncertainty-based method that chooses the instances with the least confidence scores of the predictive distribution from the unlabeled pool \cite{settles_active}. \textbf{CoreMSE} is a pool-based AL method proposed by \citep{tan2021diversity}. The method involves both diversity and uncertainty measures via the sampling strategy. It selects the diverse samples with the highest uncertainty scores from the unlabeled pool, and the uncertainty scores are estimated by the reduction in classification error. These diverse samples can provide more information to train the model to achieve high performance.

\vspace{-3mm}

\subsection{Study Setup}

% \begin{table*}[!t]
%   \caption{Datasets and the used language model}
%   \label{tab:table3}
%   \centering\small
%   \vspace{1mm}
% \renewcommand{\arraystretch}{1.25}
% \resizebox{0.6\textwidth}{!}{%
% \begin{tabular}{ccccc}
% \hline
% \textbf{\#{Class}} & \textbf{Labelled size} & \textbf{Test size} & \textbf{Lang.\ Model} & \textbf{Initial labelled size} \\ \hline
% 31              & 3763                   & 476                & ELECTRA              & 50                             \\ \hline
% \end{tabular}
% }
% \vspace{-5mm}
% \end{table*}

In the AL training process, the DA classifier was initially fed with 50 training samples randomly selected from the annotated dataset. For each AL method, the training process was repeated six times using different random seeds. A batch size of 50 was specified. We set the maximum sequence length to 128 and fine-tuned it for 30 epochs. The AdamW optimizer was used with the learning rate of 2e-5 to optimize the training of the classifier. All experiments were implemented on RTX 3090 and Intel Core i9 CPU processors.

% The dynamics validation split \cite{tan2021diversity} technique has been applied for the budget control during the training process.  Since the CoreMSE requires the ensemble technique for generating the predictive distributions, we use the Monte Carlo Dropout method to approximate the model inferences for the ensembles

% More detailed experimental settings are given in Appendix B

% \subsection{Study Experiments}

% \smallskip

% \smallskip

\section{Results}

\subsection{Estimation of Sample Informativeness}
\label{RQ1_results}
\vspace{-3mm}

\begin{figure}
\centering
\includegraphics[width=0.80\textwidth]{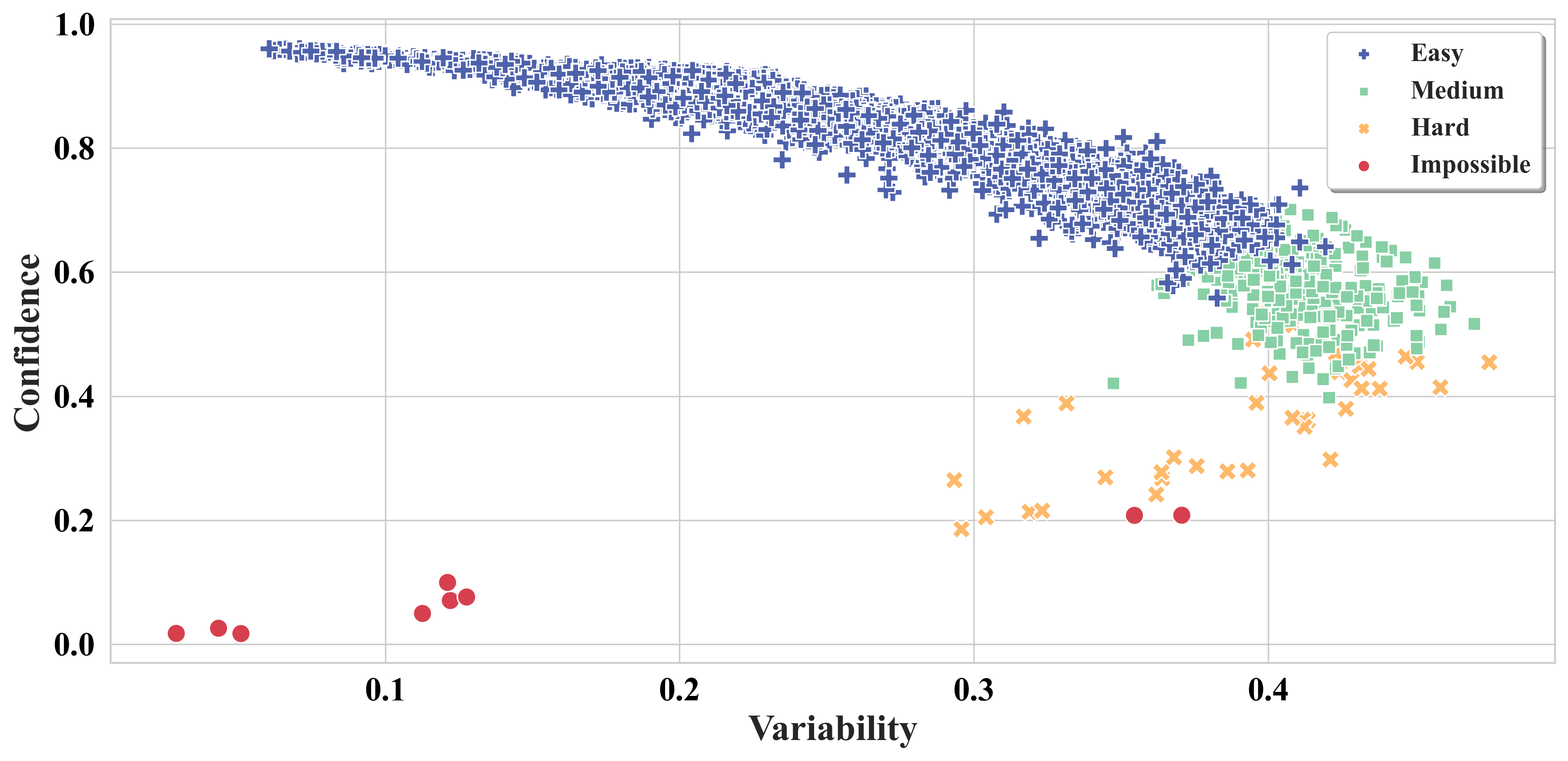}
\caption{The sample informativeness on the annotated sentences.} \label{fig1}
\vspace{-5mm}
\end{figure}

To estimate the sample informativeness level of each training instance, we employed the \textit{Data Maps}.  In Fig. \ref{fig1}, the y-axis represents the \textbf{Confidence} measure, the x-axis represents \textbf{Variability}, and the colours of instances represent \textbf{Correctness}, which are detailed in Sec. \ref{data_maps}. The samples in the \textit{Easy} group (\textit{i.e.,} blue scatters) always presented a high confidence level, which indicates that most sample points could be easily classified by our DA classier. Then, the samples in the \textit{Medium} and \textit{Hard} groups presented generally lower confidence and higher variability compared with the \textit{Easy} group, which indicates that the samples in \textit{Medium} and \textit{Hard} were more informative than those in \textit{Easy}. Lastly, the samples in the \textit{Impossible} group always presented low confidence and variability. According to \cite{swayamdipta2020dataset}, the samples in \textit{Impossible} could be the reasons for mis-annotation or insufficient training samples (\textit{e.g.,} insufficient training samples for a DA can impede the classifier's ability to capture its classification pattern) for the DA classifier. Most training samples were considered \textit{Easy} to be classified, and only a small number of samples are \textit{Impossible}. The distribution of \textit{Easy} and \textit{Impossible} points in Fig. \ref{fig1} indicates that our dataset presented high-quality annotation, which is generally considered learnable for the DA classifier.

% \begin{figure}
% \includegraphics[width=\textwidth]{Distribution of difficult_tutor.png}
% \caption{Distribution of informativeness level for each sample (Tutor) (RQ1)} \label{fig2:tutor}
% \end{figure}

% \begin{figure}
% \includegraphics[width=\textwidth]{Distribution of difficult_both.png}
% \caption{Distribution of informativeness level for each sample (Both student and tutor) (RQ1)} \label{fig2:both}
% \end{figure}

\begin{figure}
\centering
\includegraphics[width=0.80\textwidth]{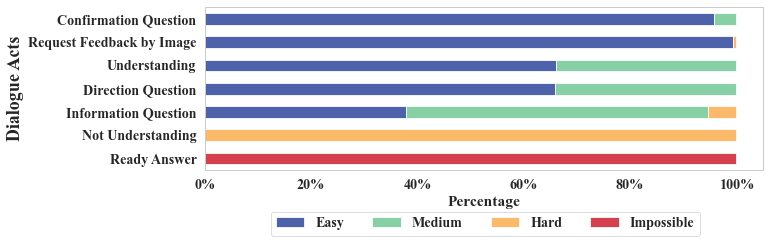}
\caption{Distribution of informativeness level for each DA (Student only)} \label{fig2:student}
\vspace{-5mm}
\end{figure}

We further investigated the distribution of informativeness level for the annotated DAs\footnote{Due to the space limit, we only present a part of the analysis results, and the full results can be accessed at \url{https://github.com/jionghaolin/INFO}.}. As described in Sec. \ref{sec:methods:subsec:da_scheme}, we mainly present the distribution of the informativeness level for each DA in Fig. \ref{fig2:student}. We sorted the DAs based on their frequency in Table \ref{tab:dialogue_act_scheme}. We observed that most samples in the \texttt{Confirmation Question} and \texttt{Request Feedback by Image} DAs were in the \textit{Easy} group, which indicates that both DAs are easy for the DA classifier to learn. Then, in the middle of Fig.\ref{fig2:student}, the \texttt{Understanding}, \texttt{Direction Question}, and \texttt{Information Question} DAs had roughly 1\% frequency in Table \ref{tab:dialogue_act_scheme}. Compared with the first two DAs in Fig. \ref{fig2:student}, the middle three DAs had a higher percentage of \textit{Medium}, which indicates that the DA classifier might be less confident in classifying these DAs. Lastly, the DAs \texttt{Not Understanding} and \texttt{Ready Answer} had the frequency lowered than 1\% in Table \ref{tab:dialogue_act_scheme}. The samples in \texttt{Not Understanding} and \texttt{Ready Answer} were considered \textit{Hard} and \textit{Impossible} to be classified, respectively. The reasons might be that the annotated samples of \texttt{Not Understanding} and \texttt{Ready Answer} in our dataset were insufficient for the DA classifier to learn the patterns. 

\vspace{-2mm}

% Therefore, it is necessary to identify these hard to learn DA instances in our unlabelled data pool to enhance the prediction performance of DA classifier. A promising solution to this necessity is to employ the Active Learning (AL) methods to automatically select the most informative instances that contribute to the classifier training process.

% To enhance the DA classifier prediction performance, it is worthwhile to further select more sample points in the \textbf{Medium} and \textbf{Hard} groups.

\subsection{Efficacy of Statistical Active Learning Methods}

\vspace{-2mm}
\begin{figure}
\centering
\includegraphics[width=0.85\textwidth]{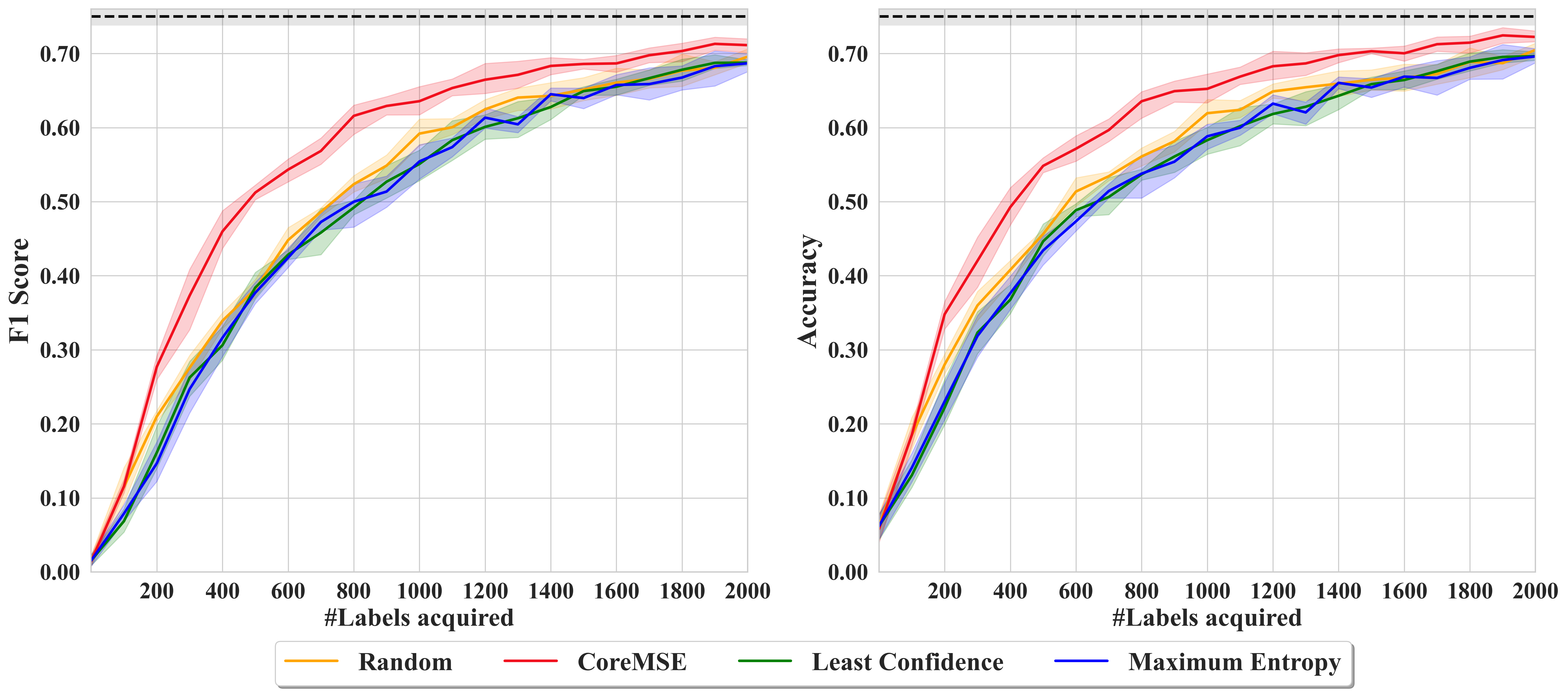}
\caption{Learning Curve of AL methods. The sampling batch for each AL method is 50. The classification performance was measured by F1 score and accuracy.} \label{fig3}
\end{figure}

To investigate how AL methods select different informative samples for training the DA classifier, we first evaluated the overall performance of the DA classifier with the support of AL methods. In Fig. \ref{fig3}, the x-axis represents the training sample size after selecting samples and the y-axis represents the classification performance measured by F1 score and classification accuracy. We compared the CoreMSE method to the baseline methods, \textit{i.e.,} Maximum Entropy (ME), Least Confidence (LC), and Random. The results in Fig. \ref{fig3} demonstrate the learning curves of the models where the CoreMSE method could help the DA classifier achieve better performance with fewer training samples compared with the baseline methods. For example, when acquiring 600 samples from the annotated data pool, the CoreMSE method could achieve roughly the F1 score of 0.55, which was equivalent to the classifier performance training on 900 samples with the use of baseline methods. It indicates that the CoreMSE method could save 30\% human annotation costs compared to the baseline methods. Whereas the efficacy of LC and ME methods was similar to that of the random baseline in both F1 score and accuracy value; this indicates that the traditional uncertainty-based AL methods (\textit{i.e.,} LC and ME) might not be effective on our classification task.

\begin{figure}
\includegraphics[width=0.95\textwidth]{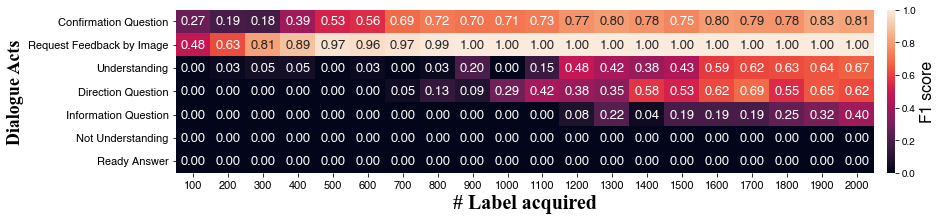}
\vspace{-2mm}
\caption{F1 score for the dialogue acts only specific to students (\textbf{Random})} \label{fig4}
\vspace{-12mm}
\end{figure}

\begin{figure}
\includegraphics[width=0.95\textwidth]{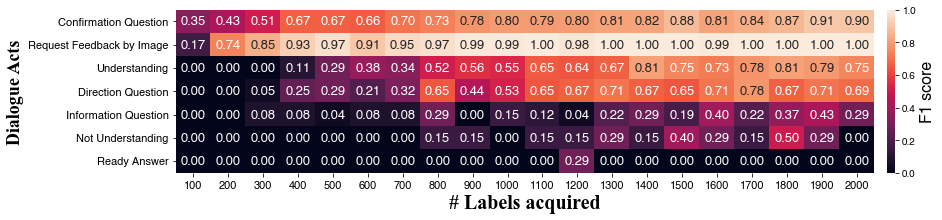}
\vspace{-2mm}
\caption{F1 score for the dialogue acts only specific to students (\textbf{CoreMSE})} \label{fig5}
\vspace{-3mm}
\end{figure}

Based on the results in Fig. \ref{fig3}, we observed that the random baseline performed slightly better than the ME and LC methods. In the further analysis of each DA, we decided to compare the efficacy between CoreMSE and the random baseline\footnote{Due to the space limit, we documented our full results in a digital appendix, which is accessible via \url{https://github.com/jionghaolin/INFO}.}. As shown in Fig. \ref{fig4} and Fig. \ref{fig5}, we sorted DAs based on their frequency in the dataset. Fig. \ref{fig4} shows the changes of F1 scores for each DA with the use of the random baseline; this indicates that the DA classifier performed quite well for classifying the \texttt{Confirmation Question} and \texttt{Request Feedback by Image} DAs as the training sample size increased. Regarding the \texttt{Understanding} and \texttt{Direction Question}, the DA classifier needed more than 1,600 annotated samples to achieve a decent performance. For the DAs \texttt{Information Question}, \texttt{Not Understanding}, and \texttt{Ready Answer}, the DA classifier almost failed to learn the patterns from the samples selected by the Random method. In comparison, Fig. \ref{fig5} shows the classification performance for each DA with the support of the CoreMSE method. The results indicate that CoreMSE could support the DA classifier to make accurate predictions for the DAs (e.g, \texttt{Understanding}) when acquiring fewer annotation samples than the random baseline. Then, though the classification performance for the DAs \texttt{Information Question} and \texttt{Not Understanding} was not sufficient, CoreMSE demonstrated the potential to improve the accuracy for both DAs as more labels were acquired. Lastly, both Random and CoreMSE methods failed to support the DA classifier to make an accurate prediction on the \texttt{Ready Answer} DA. The reason might be the fact that the \texttt{Ready Answer} DA was rare in our annotated dataset. 

% also achieved promising performance for the \texttt{Confirmation Question} and \texttt{Request Feedback by Image} DAs. Then,

% For example, with the use of CoreMSE AL method, the DA classifier could achieve 0.65 F1 score for both DAs \texttt{Understanding} and \texttt{Direction Question} at 1,100 acquired samples. whereas, the Random baseline need 2,000 samples to achieve the comparable results.

\begin{figure}
\includegraphics[width=0.95\textwidth]{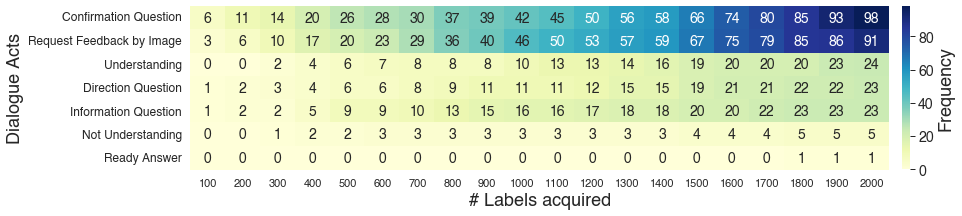}
\vspace{-2mm}
\caption{The distribution of sampling frequency for each dialogue acts (\textbf{Random})} \label{fig6}
% \vspace{-2mm}
\vspace{-12mm}
\end{figure}

\begin{figure}
\includegraphics[width=0.95\textwidth]{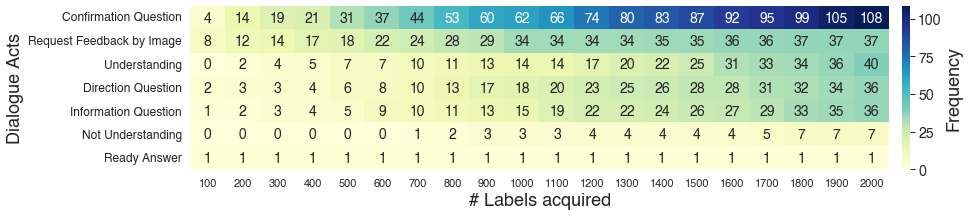}
\vspace{-2mm}
\caption{The distribution of sampling frequency for each dialogue acts (\textbf{CoreMSE})} \label{fig7}
\vspace{-2mm}
\end{figure}
% \vspace{-12mm}

% After examining the classification performance for each students' DA, it is worthwhile to know more details how the effective AL methods saved the labelling budget.

Next, we investigated the sampling process between the CoreMSE and random methods to learn more details about how the CoreMSE method saved the annotation budget. For each sampling batch, we counted the cumulative frequency for each DA in Figs.\ref{fig6} and \ref{fig7}. For example, in the first 200 acquired annotations (Fig.\ref{fig6}), the random method acquired 11 \texttt{Confirmation Question}, among which 6 of them were from the first 100 samples and 5 from the subsequent 100 acquired labels. We observed that compared with the random baseline (Fig.\ref{fig6}), the CoreMSE method (Fig.\ref{fig7}) gradually reduced sampling \texttt{Request Feedback by Image} instances from 700 labels acquired. This result indicates that the random baseline retained sampling the DA \texttt{Request Feedback by Image} instances even the F1 score achieved satisfactory performance, which might consume the budget for the human manual annotation, whereas, the CoreMSE method could alleviate the manual annotation cost when the DA classifier sufficiently learned the patterns. Then, compared with the random baseline (Fig. \ref{fig6}), the CoreMSE method (Fig. \ref{fig7}) generally selected more instances of \texttt{Understanding}, \texttt{Direction Question}, and \texttt{Information Question} DAs for training the DA classifier across the sampling process, which could explain the reason why the CoreMSE method supported the classifier achieving better performance than the random baseline.  

% \begin{figure}
% \includegraphics[width=\textwidth]{change of difficulty for each DA.png}
% \caption{Changes of distribution for each Dialogue Act (RQ2)} \label{fig4}
% \end{figure}

% \begin{figure}
% \includegraphics[width=\textwidth]{lctopk.png}
% \caption{Changes of recall for each Dialogue Act (least confidence, student only)} \label{fig5}
% \end{figure}

\vspace{-3mm}

% the dialogue act (DA) classifier can easily learn the patterns from most training instances. It indicates that not all samples contribute equally to the model performance and researchers should

\section{Conclusion}
% \vspace{-3mm}
Our study demonstrated the potential value of using a well-establish framework \textit{Data Maps} \cite{swayamdipta2020dataset} to evaluate the informativeness of instances (\textit{i.e.,} the tutorial sentences annotated with dialogue acts) in automatic classification of educational DAs. We found that most instances presented low informativeness in the training dataset, which was easy-to-learn for the dialogue act (DA) classifier. To improve the generalizability of the DA classifier on the unseen instances, we proposed that the classifier should be trained on the samples with high informativeness. Since the annotation of educational DA is extremely time-consuming and cost-demanding \cite{nye2015automated, lin2022good, rus2017analysis}, we suggest avoiding the use of random sampling for annotation. Our study provided evidence of how the state-of-the-art statistical AL methods (\textit{e.g.,} CoreMSE) could select informative instances for training the DA classifier and gradually reduce selecting the easy-to-learn instances, which can alleviate the cost of manual annotation. We acknowledged that some instances were quite rare in our original annotation, so we plan to employ the AL methods to select more of these rare instances for annotation in future research. Lastly, the AL methods might also be useful for costly evaluation tasks in the education field (\textit{e.g.,} automated classroom observation), which requires the educational experts to annotate the characteristics of behaviours among the students and teachers. Thus, a possible extension of the current work would be to develop an annotation dashboard for human experts to be used in broader educational research.

% Lastly, the current study will be further extended to develop a dashboard for human experts to annotate the DAs or the labels in other educational research (\textit{e.g.,} automated classroom observation which required the educational experts to label the characteristics of behaviors among the students and teachers)

% inspired by the previous literature \citep{ramakrishnan2021toward}, a possible extension of the current study is to employ the AL methods for supporting the automated 
% might be incredibly useful particularly for costly evaluation such as . Thus, it is worthwhile to incorporate the AL methods in the evaluation system to further boost the effectiveness. 

% Finally, we plan to extend 

\vspace{-3mm}

\bibliographystyle{splncs04}
\bibliography{mybibliography}

\end{document}